\documentclass[runningheads,a4paper]{llncs}

\usepackage{amssymb} 
\usepackage{graphicx}
\usepackage{float}

\usepackage{url}
\urldef{\mailsa}\path|{alfred.hofmann, ursula.barth, ingrid.haas, frank.holzwarth,|
\urldef{\mailsb}\path|anna.kramer, leonie.kunz, christine.reiss, nicole.sator,|
\urldef{\mailsc}\path|erika.siebert-cole, peter.strasser, lncs}@springer.com|    
\newcommand{\keywords}[1]{\par\addvspace\baselineskip
\noindent\keywordname\enspace\ignorespaces#1}

\floatstyle{ruled}
\newfloat{algorithm}{tbp}{loa}
\providecommand{\algorithmname}{Algorithm}

\floatname{algorithm}{\protect\algorithmname}
\usepackage{amsmath,graphicx}
\usepackage{epsfig}
\usepackage{graphics} 
\usepackage{epsfig} 
\usepackage{amsmath}
\usepackage{url}
\usepackage{tablefootnote}
\usepackage{subfig}
\usepackage{upgreek}
\usepackage{float}
\usepackage{pbox}

\floatstyle{ruled}
\newfloat{algorithm}{tbp}{loa}
\providecommand{\algorithmname}{Algorithm}
\floatname{algorithm}{\protect\algorithmname}

\begin{document}

\mainmatter  

\title{Uncertainty guided semi-supervised segmentation of retinal layers in OCT images}

\titlerunning{Uncertainty guided semi-supervised segmentation of retinal layers in OCT images}

%
%
%


\author{Suman Sedai \inst{1}, Bhavna Antony\inst{1},  Ravneet Rai\inst{2},   Katie Jones\inst{2}, Hiroshi ishikawa\inst{2},   Joel Schuman\inst{2},     Wollstein Gadi\inst{2} and Rahil Garnavi\inst{1}}

\institute{IBM Research - Australia, Melbourne, VIC, Australia\\ 	\email{ssedai@au1.ibm.com} \\
	 \and NYU Langone Eye Center, NYU School of Medicine, New York, NY
}

%
%

\toctitle{Lecture Notes in Computer Science}
\tocauthor{Authors' Instructions}
\maketitle

\begin{abstract}
 Deep convolutional neural networks have shown outstanding performance in medical image segmentation tasks. The usual problem when training supervised deep learning methods is the lack of labeled data which is time-consuming and costly to obtain. In this paper, we propose a novel uncertainty guided semi-supervised learning based on student-teacher approach for training the segmentation network using limited labeled samples and large number of unlabeled images. First, a teacher segmentation model is trained from the labeled samples using Bayesian deep learning. The trained model is used to generate soft segmentation labels and  uncertainty map for the unlabeled set. The student model is then updated using the softly segmented samples and the corresponding pixel-wise  confidence of the segmentation quality estimated from the uncertainty of the teacher model using a newly designed loss function. Experimental results on a retinal layer segmentation task show that the proposed method improves the segmentation performance in comparison to the fully supervised approach and is on par with the expert annotator. The proposed semi-supervised segmentation framework is a key contribution and applicable for biomedical image segmentation across various imaging modalities where access to annotated medical images is challenging. 
 
\end{abstract}

\section{Introduction}
\label{sec:intro}

Segmentation of anatomical regions in biomedical images such as optical coherence tomography (OCT) retinal scans is of great clinical significance especially for disease diagnosis, progression analysis and treatment planning. For example, the progressive thinning of circumpapillary retinal nerve fiber layer (cpRNFL) thickness measured by OCT can be used to predict the visual functional loss in patient with glaucoma \cite{pmid19684001}. 

In the past few years, Convolutional Neural Networks (CNNs) based methods such as Unet \cite{Ronneberger2015} and  Dense-Unet \cite{Simon2017,Li2018a} have achieved remarkable performance gain in medical image and natural image segmentation. The networks are trained end-to-end, pixels-to-pixels on semantic segmentation exceeded the most state-of-the-art methods without further machinery. For example, such network have been used  for retinal structure segmentation in fundus \cite{Maninis2016} and OCT  \cite{Roy2017,SedaiOMIA18} images.  Such  fully supervised  segmentation algorithms  requires large number of annotated images  to achieve reasonable robustness and accuracy.  However, acquiring pixel-wise ground truth annotations can be time-consuming and costly in medical imaging domain where only experts can provide reliable annotations. The under-supply of the labeled data motivates the need for effective machine learning methods that require limited supervision, such as semi-supervised learning.    

Semi-supervised learning tackle this problem by leveraging large number of readily available unlabeled data along with the limited labeled data to improve the performance. For example, semi-supervised approaches have been applied to different medical imaging tasks such as MRI segmentation \cite{Bai2017}, lung nodule detection and retinal vessel segmentation\cite{Xinge2011}. In another approaches, \cite{Baur2017} uses auxiliary manifold learning in the latent space for  MS lesion segmentation task and \cite{Sedai2017MICCAI} uses the feature embedding obtained from unlabeled images to segment optic cup in retinal fundus images. 

In this paper, we propose a novel semi-supervised approach to leverage unlabeled images to segment the retinal layers in OCT images. The proposed method consists of two components: (a) student segmentation network which is responsible for learning suitable data representation and learning the main segmentation task and  (b) teacher segmentation network which controls the learning of the student network by modeling the unreliability in segmentation prediction. First, the teacher model is trained on the labeled set using Bayesian deep learning to capture the uncertainty map and is used to generate \emph{soft segmentation labels} for the unlabeled samples. The uncertainty map indicates pixel-wise unreliability of the \emph{soft labels}.  Based on the uncertainty map, we further propose a  novel loss function to guide the student model by adaptively weighting regions with unreliable \emph{soft labels} to improve final segmentation performance.  Our proposed algorithm has been applied to the task of retinal layer segmentation in OCT images from the optic nerve head. Experimental results indicate that our proposed algorithm can improve the segmentation accuracy  compared to the state-of-the-art fully supervised OCT segmentation methods  and is in par with the human expert.

\section{Proposed Semi-supervised Segmentation Method} 
\begin{figure}
	\centering \includegraphics[scale=0.24]{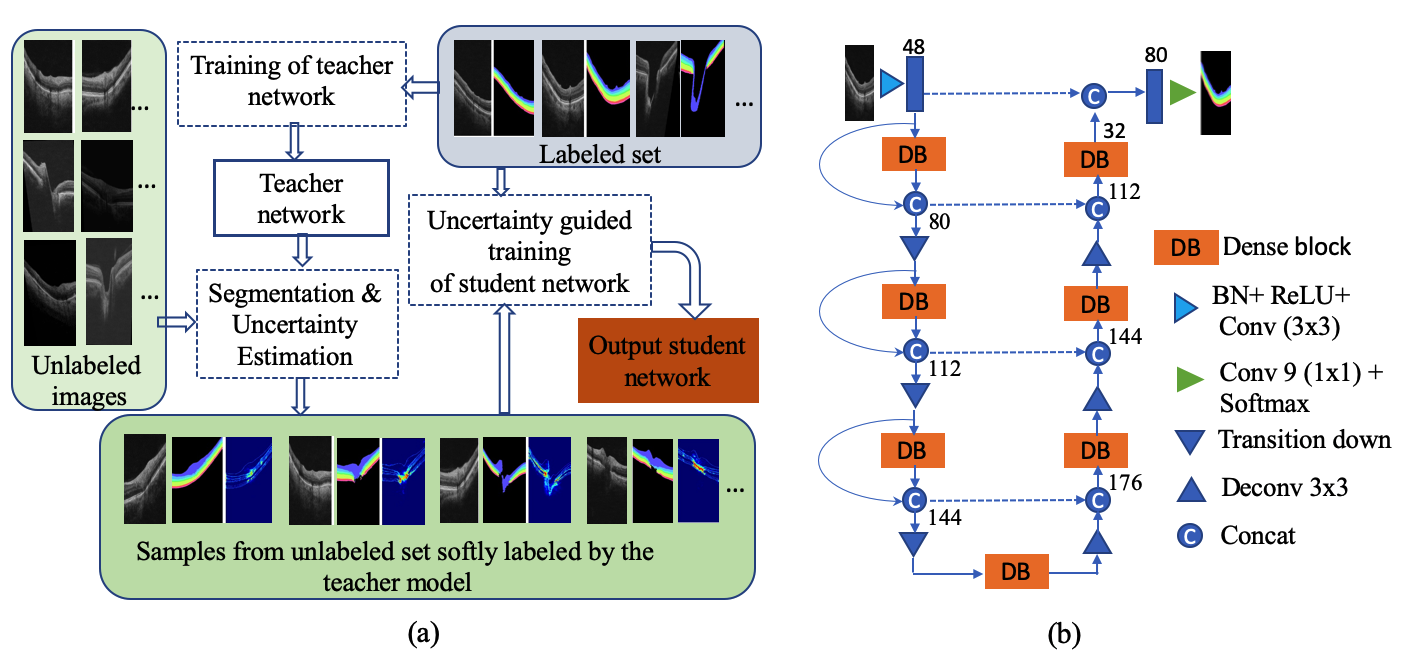} 
	\caption{ (a) Proposed semi-supervised segmentation method  based on Bayesian student-teacher learning.(b) Overview of the Dense-Unet architecture used by the teacher and the student models.}
	\label{fig:sys_diagram}
\end{figure}

In this section, we describe our proposed uncertainty guided semi-supervised learning. We assume that we are given a large set of unlabeled images $D_u=(\mathbf{x}_i)$ and a small set of high quality labeled images $D_l=\left\{ (\mathbf{x}_{i},\mathbf{y}_{i})\right\} $ where $\mathbf{x}_i$ is image and $\mathbf{y}_i$ is the segmentation annotation. As shown in Figure \ref{fig:sys_diagram}(a), our proposed approach involves two kind of deep neural networks called the teacher segmentation network $F_T(\mathbf{x}) $ and student segmentation network $F_S(\mathbf{x})$. The teacher network is trained using labeled dataset based on Bayesian deep learning to output both segmentation map  as well as uncertainty map. The teacher network is then applied to each image in the unlabeled set to obtain the segmentation label and  associated uncertainty map to generate the \emph{softly labeled} samples. The uncertainty map captures the pixel-wise confidence values indicating the reliability of the segmentation output from the teacher network and it is used to guide the training of the student network learning which we describe in Section \ref{sec:student}.  

 We use  DenseUNet architecture \cite{Simon2017,Li2018a} as our  base model for both teacher and student networks. As shown in Figure \ref{fig:sys_diagram}(b) the model consists of three dense blocks in   encoder and decoder and a bottleneck dense block with Unet like skip connections between the  output of the encoder dense blocks  and input of the decoder dense blocks.  Each dense block contains  four convolution units each comprising of $8$ $(3\times3)$ convolution, batch normalization (BN) and ReLU layer where the output of each unit is fed to the subsequent ones. Therefore each dense block produces $32$ feature maps. The final prediction layer is a convolution layer with channel number equivalent to number of classes $C$ followed by a \emph{softmax} activation function. We use \emph{SpatialDropout} \cite{Tompson2015}  layer before every convolution layer with \emph{dropout rate} of  $0.2$.

\subsection{Teacher Segmentation Network as a Bayesian Model}
\label{sec:teacher}
We model the teacher segmentation network $F_T(\mathbf{x})$ using Bayesian deep learning to capture the segmentation uncertainties for the student model, estimated with respect to the labeled data. We adopt the approach introduced by \cite{Gal2016}  based on the \emph{dropout variational inference} to compute the segmentation uncertainty.  First we train the  $F_T(\mathbf{x})$  from the  labeled sample set $D_l$ using the  class weighted categorical cross entropy loss.  For segmentation and uncertainty quantification, we enable the \emph{dropout} in test phase and the output predictive distribution is obtained by performing $K$  stochastic forward passes through the network, i.e., $\mathbf{y}^{k}=F_T^k(x), k=1,\cdots, K$ where $F_T^k$  is an effective network  after the \emph{ spatial dropout} operation. In each forward pass, the fraction of convolution  feature-maps (denoted by \emph{dropout rate}) are disabled and the segmentation score is computed using only the remaining feature-maps. The segmentation score vector $\mathbf{\bar{y}}$ is obtained by averaging the $K$ samples, via \emph{monte carlo integration}:
\begin{equation}
 \mathbf{\bar{y}}  =  \frac{1}{K}\sum_{k=1}^{K} \mathbf{y}^{k}  \label{segmentation_output}
\end{equation}
The average score vector contains the probability score for each class, ie $\mathbf{\bar{y}} = [\bar{y}_1,\cdots,\bar{y}_C ] $. The overall segmentation uncertainty for each pixel can be obtained by computing  the entropy of the average probability vector:
\begin{equation}
U(\mathbf{\bar{y}})=-\sum_{c=1}^{C}\bar{y}_{c}\text{log }\bar{y}_{c}. \label{entropy} 
\end{equation}
Higher segmentation uncertainty is obtained when the  network assigns higher probabilities to different classes for different forward passes.   Conversely, for the confident predictions, network assigns higher probability to the true class for any forward passes, resulting in lower  uncertainty value.

\begin{algorithm}[t]
	\begin{enumerate}
		\item Given an unlabeled data $D_{u}=\left\{ \mathbf{x}_{1}\right\}_{i=1}^{Nu}$ 
		and the labeled set $D_{l}=\left\{ (\mathbf{x}_{i},\mathbf{y}_{i})\right\} _{i=1}^{N_{l}}$ 
		\item Train  $F_{T}(\mathbf{x})$ using  $D_{l}$  as described in Section \ref{sec:teacher}.
		\item For each iteration, until the validation loss converges: 
		\begin{enumerate}
			\item Sample a minibatch $\hat{x}_{u}$ from $D_{u}$ and $\hat{x}_{l}$ from $D_{l}$.
			\item Compute the \emph{soft labels}  $\mathbf{z}$ and uncertainty $\mathbf{u}$  for $\hat{x}_{u}$ using Equation \ref{segmentation_output} and \ref{entropy}. 
			\item Compute the confidence map from the uncertainty map using Equation \ref{lab:control}. 
            \item Compute the labeled  and unlabeled loss using Equation \ref{eq:ssl}.

			\item Update the parameter of the student model $F_{S}$  using back-propagation.
			
		\end{enumerate}
	\end{enumerate}
	\caption{Training of the proposed semi-supervised learning method. \label{enu:algorithm}}
\end{algorithm}

\subsection{Uncertainty Guided Learning of Student Network}
\label{sec:student}

Here, we describe the process of learning the student segmentation network $F_s(\mathbf{x})$  from both unlabeled and labeled data with the guidance from teacher segmentation network $F_T(\mathbf{x})$.  We first apply $F_T(\mathbf{x})$   to the unlabeled images $\mathbf{x}_u \in U$   to obtain the soft segmentation map  $\mathbf{z}$ using Equation \ref{segmentation_output} and the associated segmentation uncertainty map  $\mathbf{u}$ using Equation \ref{entropy}. The higher  values in the uncertainty map denotes the regions where generated \emph{soft labels} are likely to be incorrect and needs to be down-weighted while updating  $F_s(\mathbf{x})$. We convert the uncertainty map $\mathbf{u}$ to obtain the normalized  confidence map  as:   
\begin{equation}
\boldsymbol{\omega} = \exp \left[-\alpha \mathbf{u}  \right] \label{lab:control} 
\end{equation}
where $\alpha$ is a positive scalar hyper-parameter and the  confidence map $\boldsymbol{\omega} \in[0,1] $  provides the pixel-wise  quality of the \emph{soft labels}  produced by $F_T(\mathbf{x})$ such that  higher uncertainty values produces low quality score and vice versa.  The unlabeled loss is then formulated as the confidence weighted cross entropy as:
\begin{equation}
L_{unlab}=-\sum_{c=1}^{C}\zeta_{c}\sum_{\forall Z_{c}}\omega_{t}\log z_{c}^{t}\label{eq:unlab_loss}
\end{equation}
where
\begin{equation}
\zeta_{c}=\begin{cases}
\underset{\forall Z_{c}}{1/\sum\omega_{t}}, & \text{if $\underset{\forall Z_{c}}{\sum\omega_{t}}, >P $}\\
0 & \text{otherwise},
\end{cases}\label{eq:effective_pixels}
\end{equation} 
such that $\zeta_{c}$ weights the contribution of each class to mitigate the effect of class imbalance in \emph{soft labels} due to its confidence weights;  $Z_{c}$ denotes the pixels region of the $c^{th}$ class in the \emph{soft label} $\mathbf{z}$ and $z_{c}^{t}$ is the \emph{softmax} output from $F_s(\mathbf{x})$   for the $t^{th}$ pixel and $c^{th}$ class. The Equation \ref{eq:effective_pixels} sets $\zeta_{c}=0$, when the effective number of pixels per class $ \smash{\sum_{\forall Z_{c}}} \omega_{t} \le P$  to improve the stability of unlabeled loss  which can happen when majority of pixels of $Z_c$ are uncertain. We empirically set $P=50$ for our retinal segmentation task. Finally, semi-supervised loss is a sum of both labeled and unlabeled loss:
\begin{equation}
L_{semi\_sup} = L_{lab}+L_{unlab} \label{eq:ssl}
\end{equation}
where $L_{lab}$ is  a \emph{categorical crossentropy} computed from the labeled mini-batch samples. The  training steps of our method is shown in Algorithm \ref{enu:algorithm}.

The proposed semi-supervised loss function encourages the network to discard the pixels with inaccurate \emph{soft labels}  generated by  $F_T(\mathbf{x})$. The hyper-parameter $\alpha$ in Equation \ref{lab:control} controls the information flow from $F_T(\mathbf{x})$ to  $F_s(\mathbf{x})$.  Intuitively small $\alpha$ allows the student to blindly follow teacher, whereas the bigger $\alpha$ controls the learning by giving emphasis on the teacher's uncertainty. For example, setting $\alpha=0$ is equivalent to using all the \emph{soft labels} whereas setting $\alpha>0$ allows probabilistic selection the \emph{soft labels} that are more certain. We empirically set the value of $\alpha$ using  validation  which we describe in Section \ref{sec:exp}.

We train the student network  for 40000 iterations or until the validation loss converges using mini-batch gradient descent and the Adam optimizer with momentum and a batch size of 1 for each labeled and unlabeled component. The learning rate is set to $10^{-5}$ which is decreased by one tenth after 10000 iterations of the training.  We augment the training images and corresponding label map masks through a mirror-image reflection and random rotation within the range of $[-15, 15]$ degrees. 

\section{Experiments} 
\label{sec:exp}
The dataset  consists of  $570$  spectral-domain optical coherence tomography (OCT) optic nerve volumes acquired   using commercial OCT device (Cirrus HD-OCT; Zeiss). Each OCT volume consists of $200$ BScans of size $1024 \times 200$. We take $700$ BScans sampled from $70$  OCT volumes to create a labeled set where the ground truth has been obtained by manual annotation of the nine boundaries from eight retinal layers \cite{Lang2013}. We convert the layer boundaries to the probability map for the eight layers regions and the background region. Therefore, the number of classes is $C=9$.  
 
  \begin{figure}[b!]
 	\centering \includegraphics[scale=0.27]{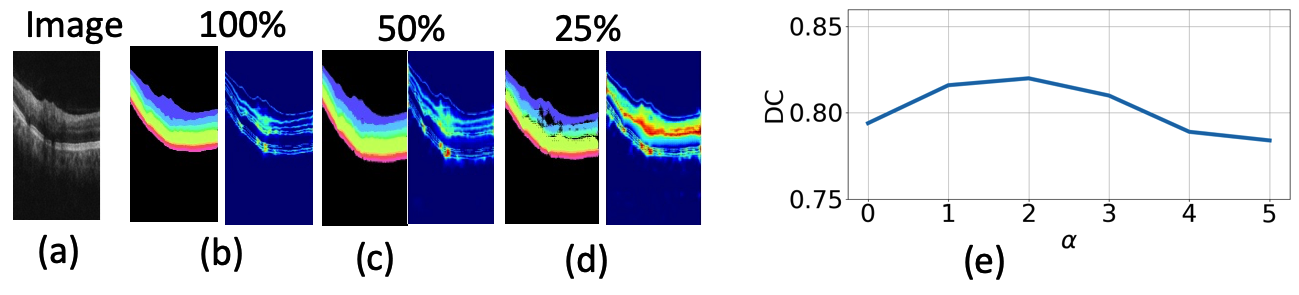} 
 	\caption{  (a)-(d) Examples of \emph{soft labels} and corresponding uncertainty map produced by the  teacher models trained using different number of labeled images.(e) The effect of different values of $\alpha$  on the performance of the student model. 
 	}
 
 	\label{fig:result_weak_labels}
 \end{figure} 
Out of $700$  labeled images, we select $490$  images from $49$ volumes to create a labeled training  set,  $70$ images from $7$ volumes for validation set and $140$ images from $14$ volume  as a test set.   Beside the annotation of expert E1 on the $700$ labeled set, we also obtained a second set of annotation for the test from the second expert E2 to compare with E1.  We then use $10000$ BScans sampled from the remaining  $500$ volumes   as an unlabeled set.

We compare our proposed uncertainty guided semi-supervised layer segmentation (U-SLS) method with the baseline fully supervised Dense-Unet (FS-DU) \cite{Simon2017}  that does not take unlabeled images into account and the plain semi-supervised learning (Plain-SL) method that blindly follows the teacher model without taking uncertainty into account, i.e., for the case where $\alpha=0$ in Equation \ref{lab:control}. Figure \ref{fig:result_weak_labels} (b)-(d) shows the examples of \emph{soft labels} and uncertainty maps produced by the teacher model $F_T(\mathbf{x})$ on an unlabeled image.  It can be seen that the teacher model trained on less number of labeled data produces less accurate \emph{soft labels} and the corresponding uncertainty map correlates with the inaccuracies in the generated \emph{soft labels}. For our method U-SLS, we set the optimal value of $\alpha=2$  using validation set as shown in Figure \ref{fig:result_weak_labels}(e).

 \begin{table}[t!]
	\caption{Retinal  segmentation performance in terms of dice coefficient of the proposed semi-supervised (U-SLS) method compared with the fully supervised  (FS-DU) and   Plain-SLS for different labeled training sizes.\label{tab:results1}}
	
	\centering{}%
	\begin{tabular}{|c|c|c|c|c|c|c|}
		\hline 
		Method  & \multicolumn{2}{c|}{\textbf{U-SLS (Proposed)}} & \multicolumn{2}{c|}{FS-DU} & \multicolumn{2}{c|}{Plain-SLS}\tabularnewline
		\hline 
		\#Images & RNFL & Average & RNFL & Average & RNFL & Average\tabularnewline
		\hline 
		490 & $0.90\pm0.04$ & $0.82\pm0.09$ & $0.88\pm0.08$ & $0.80\pm0.07$  & $0.87\pm0.08$ & $0.79\pm0.07$ \tabularnewline
		\hline 
		$240$ & $0.87\pm0.07$ & $0.80\pm0.09$  & $0.85\pm0.07$ & $0.78\pm0.12$ & $0.83\pm0.11$ & $0.76\pm0.11$\tabularnewline
		\hline 
		120  & $0.84\pm0.09$ & $0.76\pm0.12$ & $0.81\pm0.09$ & $0.74\pm0.17$ & $0.77\pm0.12$  & $0.69\pm0.17$\tabularnewline
		\hline 
	\end{tabular} 
\end{table}

Table \ref{tab:results1} compares the average Dice coefficient (DC) between the ground truth and generated segmentations by the proposed U-SLS, the plain-SLS methods and the fully supervised Dense-Unet method. The proposed method U-SLS resulted in average DC of $0.90$ for RNFL layer and $0.82$ across all eight layers when trained on the full labeled training  set and the unlabeled set improving over both FS-DU and PLain-SLS. However, when we reduce the number of labeled training samples, the improvement of the proposed method was more significant than both  FS-DU and Plain-SLS.  This demonstrates that the proposed approach improves the segmentation performance  when the number of labeled images are limited.  Moreover, the lower  performance of Plain-SLS  shows that student model is corrupted by the soft labels when  uncertainty is not taken into account. On the other hand U-SLS improves the performance by   uncertainty guided learning from the   unlabeled samples.

Table \ref{tab:res_expert_compare} compares the performance of our method with the performance of the  human annotator E2. It can be observed that performance of our method is on par with the human expert for most of the layers, including RNFl, GCL+IPL, INL, ONL and OS. We also report  the confident version of our method (U-SLS-Conf)  by evaluating on the pixels whose confidence score  $\omega > 0.5$ (computed using Equation \ref{lab:control}) which comprises  $95\%$ of the total number of pixels on average.   As expected,  U-SLS-Conf significantly improves over U-SLS which shows that the uncertainty measure produced by our method highly correlates with the segmentation inaccuracies.

\begin{figure}[t]
	\centering \includegraphics[scale=0.19]{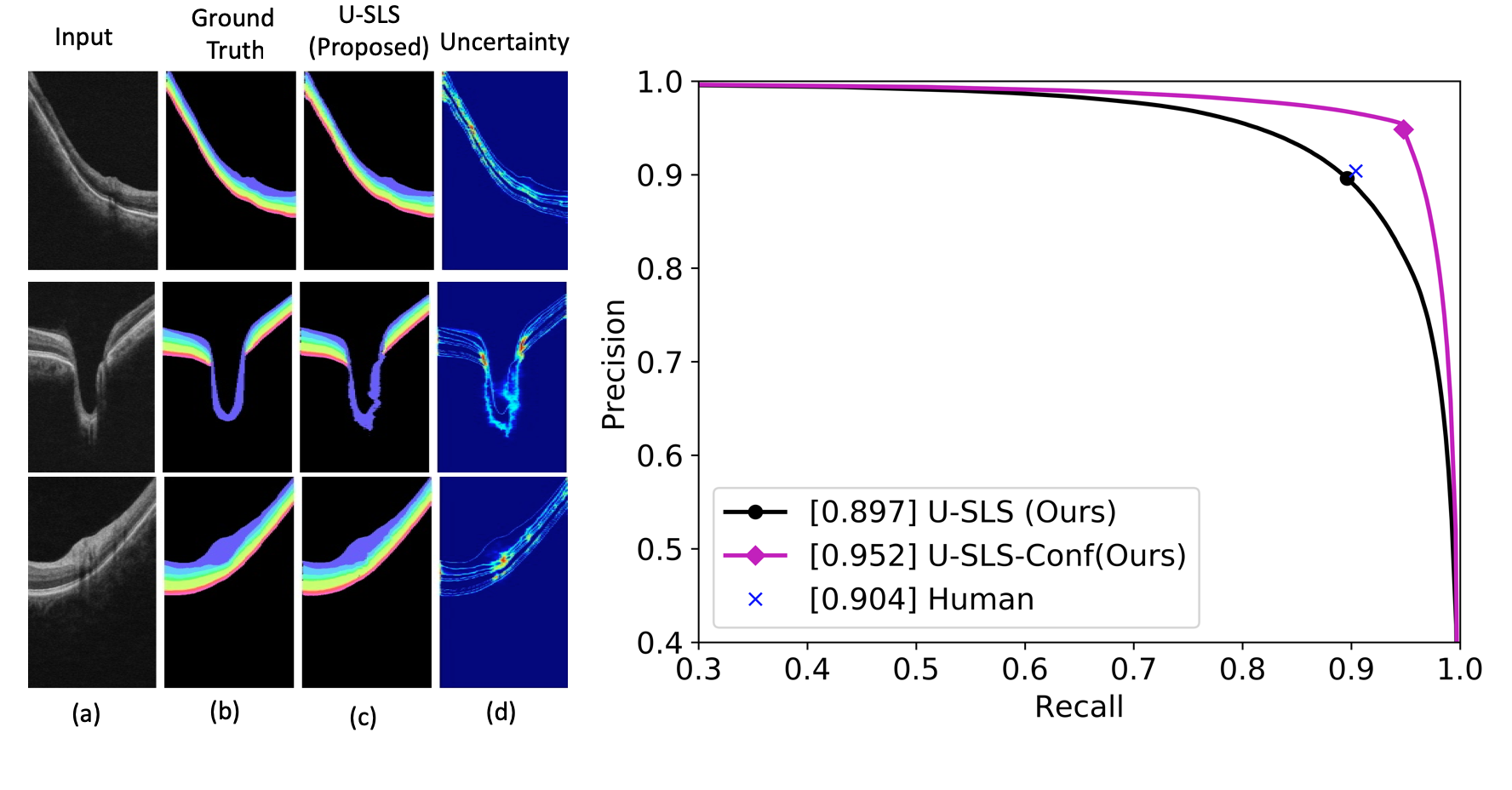}
	\caption{Example of retinal layer segmentation and uncertainty quantification using proposed U-SLS method \textbf{Left:}. (a) test images, (b) ground truth, (c) predicted segmentation map from U-SLS  (d)   uncertainty map  (warmer color denotes regions with higher uncertainty).  \textbf{Right:}  Precision recall curve comparing the proposed U-SLS method with the human expert for RNFL layer segmentation. \label{result1} 
	}
\end{figure}

Figure \ref{result1}  (left) shows the examples of the retinal layers segmentation and the generated uncertainty map.  Figure \ref{result1}  (right)  shows the precision recall curve for RNFL layer  comparing our method  with the human annotator.  This shows  that performance  U-SLS is  comparable with the human expert,   whereas the confident version, U-SLS-Conf exceeded the human expert.

\begin{table}[t]
	\caption{Performance of our proposed method compared
		with the  expert E2  }
	\label{tab:res_expert_compare}
	\centering{}%
	\begin{tabular}{|c|c|c|c|}
		\hline 
		Layer & \multicolumn{2}{c|}{Dice Coefficient} & \tabularnewline
		\hline 
		& \textbf{U-SLS (Proposed)} &E2 vs E1 & \textbf{U-SLS-Conf }\tabularnewline
		\hline 
		RNFL & $0.90\pm0.04$  & $0.90\pm0.07$ & $0.95\pm0.03$\tabularnewline
		\hline 
		GCL+IPL & $0.83\pm0.06$ & $0.84\pm0.11$ & $0.91\pm0.06$\tabularnewline
		\hline 
		INL & $0.78\pm0.07$ & $0.78\pm0.14$ & $0.90\pm0.05$\tabularnewline
		\hline 
		OPL & $0.68\pm0.09$ & $0.74\pm0.14$ & $0.79\pm0.01$\tabularnewline
		\hline 
		ONL & $0.92\pm0.03$ & $0.91\pm0.06$ & $0.97\pm0.01$\tabularnewline
		\hline 
		IS & $0.74\pm0.09$ & $0.77\pm0.14$ & $0.81\pm0.10$\tabularnewline
		\hline 
		OS & $0.82\pm0.06$ & $0.83\pm0.10$ & $0.90\pm0.04$\tabularnewline
		\hline 
		RPE & $0.82\pm0.06$  & $0.85\pm0.08$ & $0.89\pm0.001$\tabularnewline
		\hline 
		Average & $0.82\pm0.07$  & $0.84\pm0.12$ & $0.91\pm0.08$\tabularnewline
		\hline 
	\end{tabular} 
\end{table}

\section{Conclusion}

In this paper, we presented a novel and effective semi-supervised method, based on student-teacher framework, for segmentation of OCT images of retina. The proposed method is able to leverage large volume of unlabeled noisy data and incorporate uncertainty for improved segmentation of retina structures, compared to the state-of-the-art fully and semi-supervised segmentation methods. We have demonstrated that the proposed uncertainty guided method can effectively transfer knowledge from the teacher to the student model for the segmentation task and is able to generate expert-level segmentation using limited number of labeled samples. Therefore, our approach is useful in clinical applications where access to large volume of annotated images (which is needed for state-of-the-art fully supervised approaches) is challenging. Although, we have applied our approach in retinal image segmentation, we believe that our method is equally applicable to other modalities.

\bibliographystyle{splncs03}
\bibliography{refs_ssl}

\end{document}